\documentclass[letterpaper, 10 pt, conference]{ieeeconf}
\IEEEoverridecommandlockouts
\renewcommand{\paragraph}[1]{\vspace{0.5mm} \noindent \textbf{#1}}
\let\labelindent\relax

\newcommand{\sd}[1]{\vspace{0.5mm} \noindent \textit{#1}}
\newcommand{\td}[1]{\vspace{0.5mm} \noindent \textit{#1}}

\newcommand{\sBox}{\texttt{AnimalDorms}\xspace}
\newcommand{\sScene}{\texttt{SceneTwins}\xspace}
\newcommand{\sScan}{\texttt{GroceryCheckout}\xspace}

\newcommand{\tBox}{\texttt{PackBox}\xspace}
\newcommand{\tScene}{\texttt{ArrangeDesk}\xspace}
\newcommand{\tScan}{\texttt{ScanBottle}\xspace}

\usepackage{algpseudocode}
\usepackage{amsmath}
\usepackage{amssymb}
\usepackage{amsfonts}
\usepackage{array}
\usepackage{booktabs}
\usepackage{cellspace}
\usepackage{enumitem}
\usepackage{graphics}
\usepackage{epsfig}
\usepackage{longtable}
\usepackage{mathptmx}
\usepackage{marginnote}
\usepackage{multirow}
\usepackage{pdfx}
\usepackage{setspace}
\usepackage{soul}
\usepackage{tabularx}
\usepackage{tcolorbox}
\usepackage{times}
\usepackage{wrapfig}
\usepackage{xcolor}
\usepackage{xspace}
\usepackage{inconsolata}
\usepackage[cal=pxtx]{mathalpha}
\usepackage[vlined,linesnumbered,ruled,resetcount]{algorithm2e}
\usepackage[capitalise, nameinlink]{cleveref}
\usepackage[natbib=true, style=ieee, citestyle=numeric-comp]{biblatex}

\title{{\LARGE \bf RoboCade: Gamifying Robot Data Collection} \vspace{-5mm}}

\author{\cr Suvir Mirchandani$^{*1}$, Mia Tang$^{*1}$, Jiafei Duan$^{2}$, Jubayer Ibn Hamid$^{1}$, Michael Cho$^{3}$, Dorsa Sadigh$^{1}$%
\vspace{7mm}
\thanks{$^{*}$Equal contribution. \texttt{\{suvir,miatang\}@cs.stanford.edu}}%
\thanks{$^{1}$Stanford University; $^{2}$University of Washington; $^{3}$FrodoBots}%
}  

\usepackage[subtle,trackingfraction=0.99]{savetrees}
\usepackage[font=footnotesize]{caption}
\hypersetup{
    colorlinks=true,
    linkcolor=orange,
    filecolor=magenta,      
    urlcolor=orange,
    citecolor=orange,
}

\begin{document}

\maketitle
\thispagestyle{empty}
\pagestyle{empty}

\begin{abstract}
Imitation learning from human demonstrations has become a dominant approach for training autonomous robot policies. However, collecting demonstration datasets is costly: it often requires access to robots  and needs sustained effort in a tedious, long process. These factors limit the scale of data available for training policies. We aim to address this scalability challenge by involving a broader audience in a gamified data collection experience that is both accessible and motivating. Specifically, we develop a gamified remote teleoperation platform, RoboCade, to engage general users in collecting data that is beneficial for downstream policy training. To do this, we embed gamification strategies into the design of the system interface and data collection tasks. In the system interface, we include components such as visual feedback, sound effects, goal visualizations, progress bars, leaderboards, and badges. We additionally propose principles for constructing gamified tasks that have overlapping structure with useful downstream target tasks. We instantiate RoboCade on three manipulation tasks---including spatial arrangement, scanning, and insertion. To illustrate the viability of gamified robot data collection, we collect a demonstration dataset through our platform, and show that co-training robot policies with this data can improve success rate on non-gamified target tasks (+16--56\%). Further, we conduct a user study to validate that novice users find the gamified platform significantly more enjoyable than a standard non-gamified platform (+24\%). These results highlight the promise of gamified data collection as a scalable, accessible, and engaging method for collecting demonstration data. Videos are available at \url{robocade.github.io}.
\end{abstract}
\section{Introduction}
\label{sec:introduction}

In recent years, imitation learning from human demonstrations has become an increasingly popular approach for training robot policies to perform versatile skills \cite{zhao2023learning, kim2024openvla, zhao2025aloha, black2024pi_0}.
A key factor powering these approaches is investment in real-world robot data collection.
However, as modern imitation learning methods scale up, so do the requirements of data collection---often on the order of hundreds to thousands of demonstrations for individual tasks \cite{zhao2025aloha, black2024pi_0}. Thus, scaling real-world data collection remains a crucial bottleneck.

Two key challenges underlie this bottleneck: limited access to robots, and limited incentive to engage in data collection.
First, robot data collection typically requires physical access to costly robot hardware, restricting participation to a small pool of trained operators and researchers.
Second, while remote teleoperation platforms \cite{Mandlekar2018ROBOTURKAC, dass2024telemoma} and robot-free platforms \cite{duan2023ar2, chi2024universal} have emerged as a promising way to broaden access, they are typically designed with functionality as the primary concern. These systems often do not actively prioritize user engagement in their design.
Thus, even when access is no longer a limiting factor, there remains little incentive for broader participation: the data collection process remains tedious and unengaging for most users.
As a result, most datasets in academia and industry are collected in-house through dedicated data collection efforts by researchers or hired operators \cite{dasari2019robonet, rt22023arxiv, walke2023bridgedata, Khazatsky2024DROIDAL, black2024pi_0, lee2025molmoact}. 

To address the challenges of access and incentive, we propose examining robot data collection through the lens of \textit{gamification}---the integration of game-like elements into non-gaming systems \cite{deterding2011game}---specifically in the context of remote teleoperation. Gamification has been successfully applied to diverse domains from education to commerce to increase user participation and engagement \cite{koivisto2019rise}.
In this vein, we propose a framework for gamifying robot data collection by recasting remote teleoperation---typically via a purely utilitarian interface---into an engaging interactive experience. By embedding gamification principles into both system interface and task design, we aim to enable experiences that are not only engaging but also yield useful robot demonstrations.

% === Front Figure ===
\begin{figure}
    \centering
    \includegraphics[width=\linewidth]{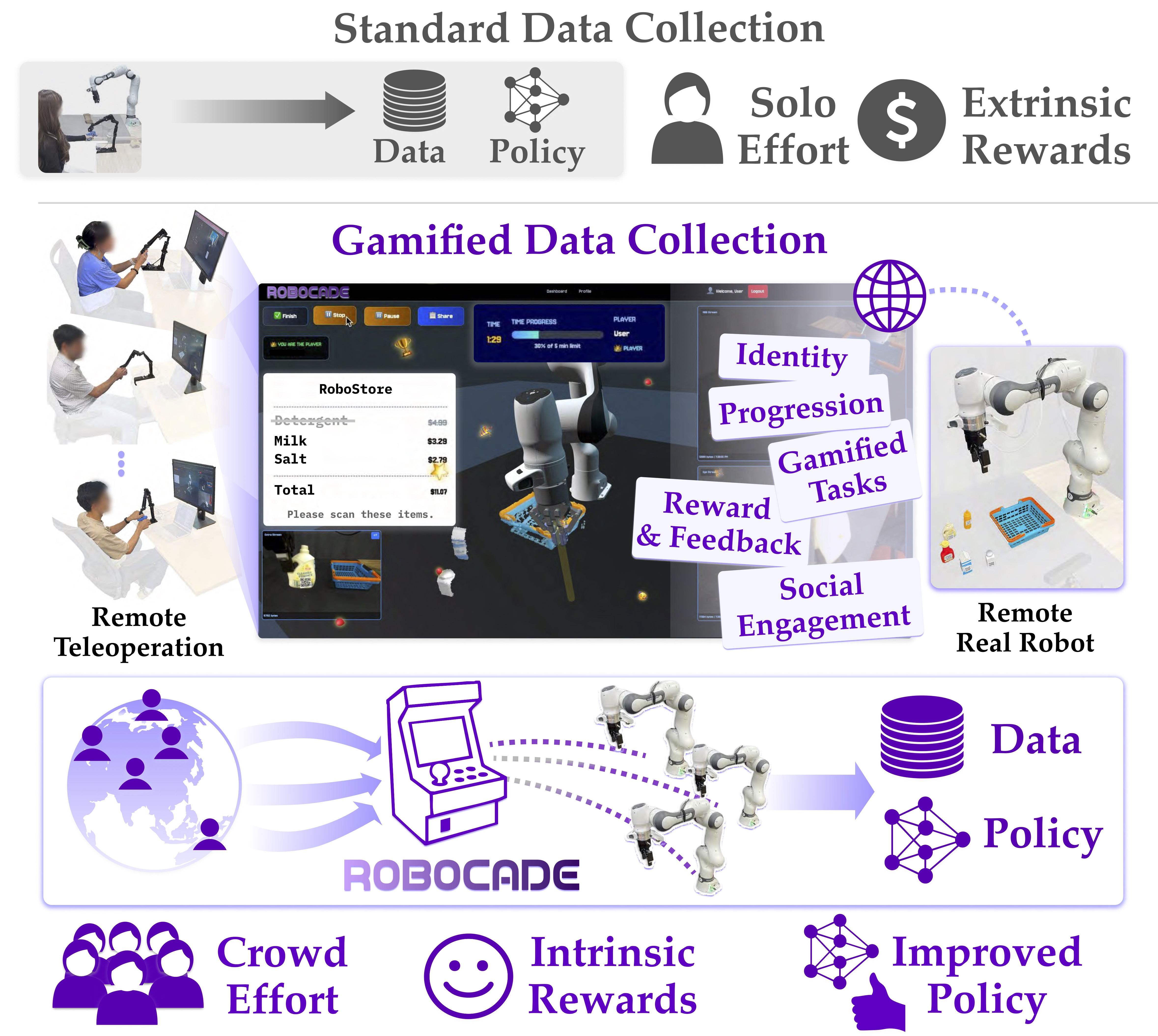}
    \caption{We develop \textbf{RoboCade}, a platform that gamifies the collection of robot demonstration datasets. In standard data collection approaches (\textit{top}), hired operators or researchers collect demonstrations using in-person teleoperation, which requires access to a robot and can be tedious and time-consuming. RoboCade (\textit{bottom}) is a remote data collection platform that integrates gamification into both system and task design, making robot data collection more engaging and accessible to a broader set of users.}
    \vspace{-1em}
    \label{fig:front}
\end{figure}

We instantiate this framework in a web-based platform, RoboCade. Built on top of a 3D-printable GELLO controller \cite{wu2024gello}, which enables intuitive teleoperation at a low-cost, RoboCade allows users to teleoperate real robots remotely, with game mechanics embedded throughout the data collection process. We instantiate this system on 3 manipulation tasks---including spatial rearrangement, insertion, and scanning tasks---that are recast as interactive game-like experiences. \looseness=-1

Our contributions are as follows.
First, we develop a {gamified remote teleoperation platform} based on a web-based interface with a GELLO controller, featuring a variety of gamified components.
Second, we develop a set of {principles for designing gamified robot manipulation tasks} that integrate game-like mechanics but retain relevance to downstream target tasks.
Third, we {co-train policies with data collected by our gamified interface}, illustrating the viability of gamification in robot data collection. Our simple co-training recipe combines demonstrations of gamified tasks collected using RoboCade and non-gamified target tasks collected using standard teleoperation. We find that co-training policies with data collected through our gamified platform improves in-distribution success on 3 non-gamified target tasks (+16--56\%) with additional boosts in out-of-distribution performance up to 20\%. Likewise, we observe benefits to including data from gamified tasks in the context of fine-tuning pre-trained vision-language-action models on target tasks. 
Finally, we demonstrate through a {human subjects study} ($N=18$) that users find our gamified interface more intuitive (+27\%), enjoyable (+24\%), and motivating (+24\%) compared to a non-gamified platform. 
\section{Related Work}
\label{sec:related_work}

\paragraph{Gamification of Data Collection.} Outside of robotics, gamification \cite{deterding2011game} is a widely studied technique to improve user engagement across domains ranging from education to commerce \cite{koivisto2019rise, hamari2014does}. Early work on Games-with-a-Purpose (GWAPs) \cite{von2008designing} proposed the idea of channeling user interactions in game-like experiences towards collecting data for machine learning tasks such as image captioning \cite{vonAhn2004labeling}, object recognition \cite{von2006peekaboom}, and commonsense reasoning \cite{von2006verbosity}.
However, the connection between gamification and robot data collection remains underexplored \cite{leite2025call}. Our work aims to bridge the gap, scaling robot data collection through gamifying the experience to involve general users.

\paragraph{Robot Data Collection Platforms.} 
A variety of interfaces exist for collecting robot demonstrations. These span four families, each with trade-offs in terms of control fidelity, safety, operator availability, and overall scalability.
%Prior work has introduced a variety of interfaces for collecting robot demonstrations, spanning four families, each with trade-offs.
Kinesthetic teaching \cite{argall2009survey} is precise and safe, but is labor-intensive and requires collocation with a robot. Common teleoperation input devices (SpaceMouse \cite{zhu2023viola}, gamepads \cite{wang2025roboeval}, commercial VR controllers \cite{rt12022arxiv}, AR \cite{duan2023ar2,wang2024eve}, smartphones \cite{Mandlekar2018ROBOTURKAC}) boost throughput and allow remote access, but can be cognitively demanding and latency-sensitive. Wearable end-effector systems (DexCap \cite{wang2024dexcap}, UMI \cite{chi2024universal}, DexUMI \cite{xu2025dexumi}) which map human hand poses to the end-effector can allow for higher dexterity and demonstration rates, but can require custom hardware, calibration, and careful retargeting. Recently, purpose-built puppeteering platforms such as ALOHA \cite{zhao2023learning} and GELLO \cite{wu2024gello} are designed to streamline data collection while preserving real-robot fidelity. %and democratize robot access.
% All methods have tradeoffs in terms of control fidelity, safety, scale, and operator availability.
Rather than introducing a new teleoperation device, we leverage GELLO \cite{wu2024gello}, which enables intuitive and high-fidelity teleoperation of real robot arms at a low-cost. Prior works typically treat the input controller for data collection in a purely utilitarian manner. In this work, we build a remote gamified data collection system using GELLO, enabling us to study user engagement in robot data collection, while maintaining real-robot fidelity and removing the requirement for proximity to a real robot.

\paragraph{Crowdsourcing Robot Data.}
Crowdsourcing has emerged as a scalable avenue for robot data collection. Early systems such as RoboTurk \cite{Mandlekar2018ROBOTURKAC} crowdsource demonstrations via a smartphone teleoperation interface, enabling remote workers to control robots. More recent efforts extend this paradigm to in-the-wild settings---for example, remotely teleoperated cars to harvest navigation trajectories \cite{hirose2025learning} and low-cost hardware that brings data collection into homes and labs (e.g., SO-100/101 robots via the LeRobot platform \cite{shukor2025smolvla}). Recent work \cite{mirchandani2025robocrowd} studies crowdsourcing of in-person robot demonstrations in public spaces, but requires deployment of physical robots, limiting scalability. Other lines of work use crowds to guide exploration in real-world RL \cite{balsells2023autonomous} or to collect interaction data through higher-level abstractions \cite{chung2014accelerating}, trading off low-level control for accessibility. A common limitation is reliance on extrinsic incentives, such as monetary or token-based payments, which can cap long-term scale and sustainability. In contrast, RoboCade explores gamification of remote data collection, harnessing aspects such as intrinsic motivation and competition, with the potential to enable scalable crowdsourcing without direct monetary incentives.
\section{RoboCade}
\label{sec:method}

% === Main Figure ===
\begin{figure*}
    \centering
    \includegraphics[width=\linewidth]{figures/overview-compressed.jpg}
    \caption{\textbf{System and Task Design Overview.} RoboCade uses a GELLO controller \cite{wu2024gello} with a web-based interface to enable remote teleoperation of real robots. We embed gamification principles into the design of the system and tasks. (\textit{Top}) Into the system, we incorporate visual and auditory feedback and rewards, challenge and goal visual cues, identity and progression, and social engagement. (\textit{Bottom}) Into the design of tasks, we incorporate narrative, goal diversity, an appropriate level of challenge, and overlapping skills with relevant downstream tasks.}
    \vspace{-1.5em}
    \label{fig:overview}
\end{figure*}

We develop RoboCade: a platform that gamifies the collection of robot demonstration datasets. In \cref{sec:method.problem_setting}, we motivate the need for an accessible and engaging data collection platform in the context of imitation learning. In \cref{sec:method.gamifying}, we reframe robot data collection as a gamified experience and establish desiderata for a platform that enables gamified data collection. Next, in \cref{sec:method.system_design} and \cref{sec:method.task_design}, we describe how we achieve these goals through a combination of system design and task design. We instantiate this platform on three tasks (\cref{sec:method.task_details}). In \cref{sec:experiments}, we validate RoboCade as a data collection platform via policy learning experiments and a human subjects study. \looseness=-1

\subsection{Problem Setting: Data Collection for Imitation Learning}
\label{sec:method.problem_setting}

Imitation learning (IL) aims to learn the behavior of an expert policy $\pi_E$ performing a task $\mathcal{\tau}$. IL is often formulated as behavior cloning, where a policy $\pi_\theta : \mathcal{O} \to \mathcal{A}$ parameterized by $\theta$ is trained using supervised learning from a dataset $\mathcal{D_\tau} = \{ \xi_1, \ldots, \xi_N \}$ composed of $N$ expert demonstrations $\xi_i$. Each expert demonstration $\xi_i$ is a sequence of observation-action transitions $\{ (o_0, a_0), \ldots, (o_{T_i}, a_{T_i}) \}$ with $o_i \in \mathcal{O}$, $a_i \in \mathcal{A}$. \looseness=-1 % Observations are typically composed of multiple RGB images and the proprioceptive state of the robot.

In practice, the costs of collecting $\mathcal{D}_\tau$ can be substantial due to the human effort involved. These datasets are often collected through long and tedious teleoperation sessions in which operators repeatedly collect demonstrations for specific tasks. As a result, the size and composition of $\mathcal{D_\tau}$ is closely tied to access to operators, availability of robot hardware, and any external incentive (e.g., monetary payment) for collecting data. In this work, we develop a platform that gamifies robot data collection, wherein a broader population can be engaged in data collection through a combination of a gamified remote teleoperation interface and design of engaging tasks.

%%%%%%%%%%%%%%%%%%%%%%%%%%%%%%%%%%%%
\subsection{Gamifying Robot Data Collection}
\label{sec:method.gamifying}

While prior work has demonstrated the effectiveness of gamification across domains~\cite{koivisto2019rise}, applying it to robot data collection introduces unique challenges.
First, the system must foster intrinsic motivation through engaging gameplay, without relying on users to care about the data itself, broadening the pool of potential contributors.
Second, it must support smooth remote teleoperation while concurrently streaming gamified interface elements with minimal latency.
Finally, the system should leverage low-cost teleoperation controllers to enable deployment beyond research labs and increase accessibility for a broader audience. From these considerations, we derive three goals for our system:

\begin{itemize}[leftmargin=*]
    \item \textit{Engaging.} The system should provide a gamified experience that is intuitive, engaging, and motivating for users. \looseness=-1
    \item \textit{Remote-friendly.} The system should support low-latency, smooth teleoperation for users across diverse locations.
    \item \textit{Low-cost.} The system should be inexpensive to build and maintain, enabling users to access with minimal resources.
\end{itemize}

\paragraph{Operators as Players.} We reconceptualize the role of robot operators by framing them as \textit{players}. Rather than treating teleoperation as a purely functional means of issuing robot commands to collect demonstration data, our system invites users to engage as players in a gamified experience---participating for gameplay rather than solely for data collection. This reframing translates manipulation tasks into interactive challenges. This lowers barriers for broader participation by enabling playful goal-driven interaction, reducing the need for external incentives.

\paragraph{Episodes as Attempts.} We frame each teleoperation episode as an \textit{attempt} within a game-like experience. This framing situates demonstrations in a familiar cycle of trial and retry, where users are encouraged to make progress across multiple attempts rather than viewing a single episode as final. 

\paragraph{Teleoperation as Gameplay.}  We frame teleoperation as a gameplay experience by incorporating game design elements and curating tasks with gamified structures. Next, we detail the components through which this transformation is realized. \looseness=-1

%%%%%%%%%%%%%%%%%%%%%%%%%%%%%%%%%%%%
\subsection{System Design: Gamified Remote Teleoperation}
\label{sec:method.system_design}

We integrate a specific set of game design elements guided by four core design principles,  aimed at fostering engagement from a general audience and, in turn, enabling more scalable robot data collection. \looseness=-1

% D1  \textit{Feedback \& Reward}, D2 \textit{Challenge \& Goal Structure}, D3 \textit{Identity \& Progression}, and D4 \textit{Social Engagement}.  Below we expand on specific instantiations for each design principle.

% \begin{enumerate}[label=D\arabic*]
%     \item \textit{Feedback \& Reward.} The system should provide timely and informative feedback on player actions, reinforcing progress and performance through clear visual, auditory, or symbolic cues.  
%     \item \textit{Challenge \& Goal Structure.} The experience should balance difficulty and attainability by setting explicit goals and calibrated time limits, encouraging persistence while avoiding frustration.  
%     \item \textit{Identity \& Progression.} The system should establish a sense of personal identity for each user and support progression over time through mechanisms such as profiles, points, or badges.  
%     \item \textit{Social Engagement.} The system should create opportunities for social connection, allowing players to observe, compare, and compete with others in ways that sustain community participation.  
% \end{enumerate}

\sd{SD1. Feedback \& Reward.} 
Timely feedback on user actions is crucial to engagement \cite{sweetser2005gameflow}. We incorporate several forms of feedback to reinforce progress---for example, visually highlighting the robot gripper in the interface as a confirmation of a grasp, and rewarding task completion with celebratory visual and auditory effects (\cref{fig:overview}). 

\sd{SD2. Challenge \& Goal Cues.}  
The system introduces visual cues for challenges and goals to sustain engagement. Each task is bounded by a predefined time limit that is appropriately calibrated yet presents a challenge for non-expert users. This time limit is visualized with a progress bar that fills as time elapses (\cref{fig:overview}). Task goals are visualized in ways tailored to the task itself. For instance, in our scanning task, the object list which the user must complete is presented as a grocery store receipt (\cref{fig:overview}). In our rearrangement task, a scene that the user must recreate with objects is virtually projected onto the table surface. \looseness=-1

\sd{SD3. Identity \& Progression.}  
Establishing a sense of personal identity within digital systems is a key driver of sustained engagement and participation \cite{username_avatar_engagement}. In our platform, users create a persistent profile with a username and avatar that is visible across sessions and displayed on every page, giving them a distinct RoboCade identity. Game participation contributes to a point system that unlocks badges as users engage in more episodes and games, providing a lightweight progression structure that can reinforce longer-term involvement.

% The system establishes personal identity and visible markers of progress. 

\sd{SD4. Social Engagement.}  
Social engagement, e.g., through competition among participants, is an important aspect of engagement in games \cite{medler2011analytics}. We incorporate a public leaderboard which ranks users by cumulative game points. In addition, each game session generates a unique shareable URL, enabling spectators to observe sessions in real-time.

\subsection{Task Design: Support Tasks for Co-Training}
\label{sec:method.task_design}

\begin{figure}
    \centering
    \vspace{1mm}
    \includegraphics[width=1\linewidth]{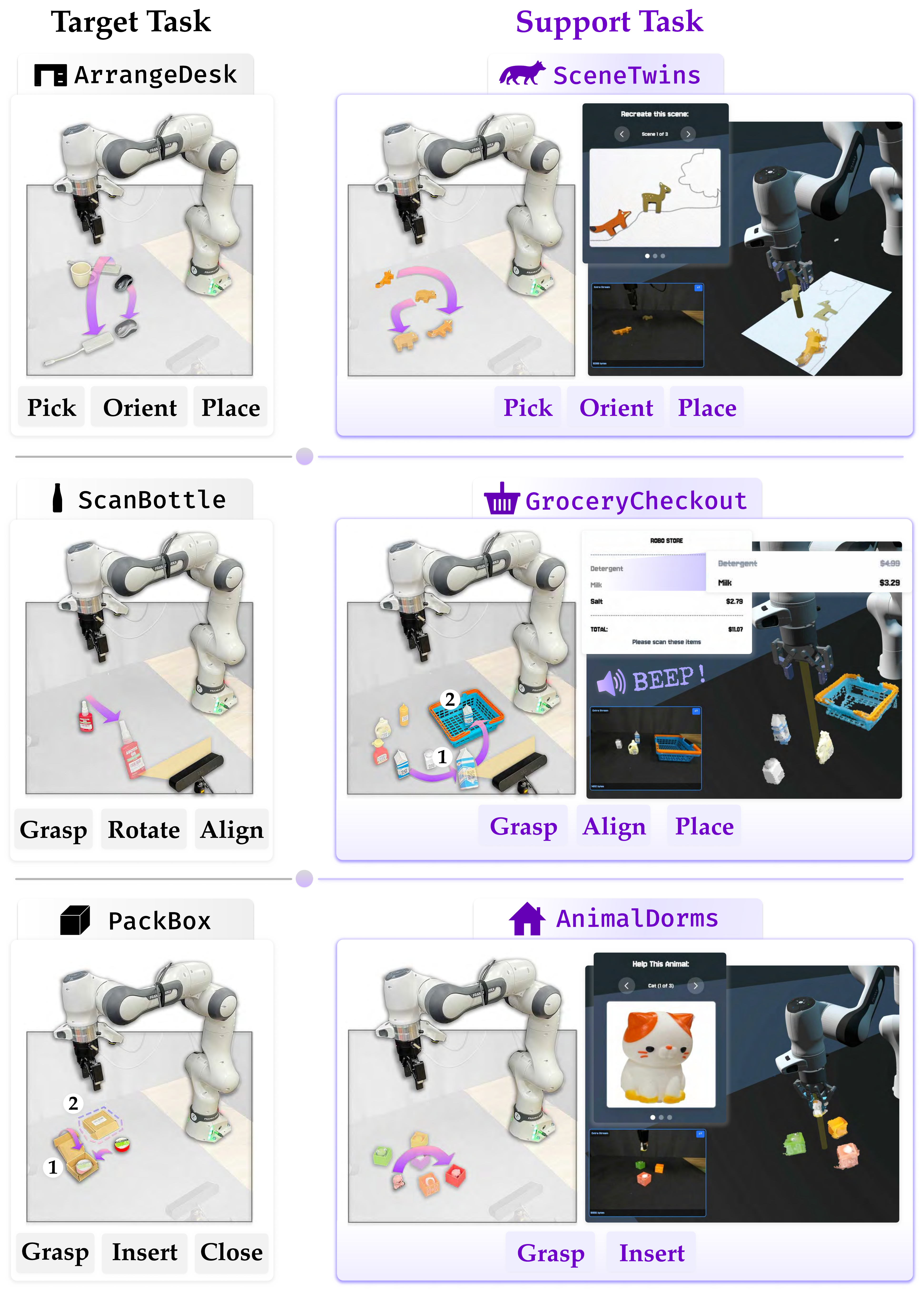}
    % \vspace{-3mm}
    \caption{\textbf{Task Design}. We show 3 pairs of support tasks (\textit{right}) and target tasks (\textit{left}). Rearrangement (\textit{top}): Both tasks involve moving 2 objects; \tScene focuses on organizing electronics, while \sScene uses animal blocks with diverse layouts and a virtual goal overlay. Scanning (\textit{middle}): both tasks involve grasping and aligning objects for scanning; \tScan requires more precision due to the object geometry and randomized orientations, while \sScan adds object variety, a grocery checkout narrative, sound effects, and a basket placement step. Insertion (\textit{bottom}): both tasks require packing, but \tBox adds complexity by closing a box lid while \sBox adds narrative and object variety.}
    \vspace{-5mm}
    % \caption{\textbf{Task Design.} We illustrate 3 instances of designing tasks for gamified data collection (\textit{right}) based on downstream target tasks (\textit{left}). In the rearrangement tasks (\textit{top}), both \tScene and \sScene require moving two objects. \tScene requires moving electronics away from a mug and lining them up at the side of a table. \sScene instead uses animal blocks initialized in more diverse locations with more diverse goal positions, indicated to the user using a virtually projected desired scene overlay. In the scanning tasks (\textit{middle}), both \tScan and \sScan require grasping objects and aligning objects with a scanning camera. Both grasping and alignment are slightly harder in \tScan, as the orientation of the bottle of threadlocker is randomized, and the thinner cap requires more precise grasping. \sScan is adds additional diversity in terms of objects, as well as a narrative of grocery checkout---users hear an auditory beep upon scanning, and there is an extra step of placing items in a grocery basket. In the insertion tasks (\textit{bottom}), both tasks require packing an object. \sBox uses a more diverse set of objects (animal toys); \tBox requires more complex manipulation as it also requires closing the lid of a cardboard box.}
    \label{fig:tasks}
\end{figure}

In this section, we propose principles for designing tasks for gamified remote data collection. Driving these principles are two main considerations: tasks should be engaging but nevertheless produce useful data for non-gamified tasks. To this end, we approach task design as a transformation from a downstream \textit{target task} $\tau$ into a gamified \textit{support task} $\tau'$. In \cref{sec:method.task_details}, we provide 3 instantiations of support tasks.
%In \cref{sec:experiments.policy_learning}, we validate the idea that support tasks can be beneficial for learning policies on target tasks, specifically through a co-training paradigm.

\td{TD1. Narrative.} Robot data collection tasks are typically specified through functional language descriptions---e.g., ``put object A at location B'' \cite{Khazatsky2024DROIDAL}. To increase immersion, we embed the same action sequence within a narrative or theme \cite{flatla2011calibration}, supported via textual and visual prompts. For example, inserting toy objects into a box can be reframed as ``helping a character find its home'' (\cref{fig:overview}), inviting users to engage with the task as part of a story. \looseness=-1

\td{TD2. Goal Diversity.} Repeating the same objective across trials can quickly lead to disengagement. We introduce variations in task goals while preserving underlying manipulation skills, transforming repetitive actions into distinct challenges. For instance, grasping skills can be practiced with different objects, or rearrangement skills can feature different spatial goals (\cref{fig:overview}), adding novelty to each episode.

\td{TD3. Challenge.} Tasks risk being either too trivial to sustain engagement or too difficult for novices to attempt. Guided by the concept of flow \cite{Csikszentmihalyi_1988}, we calibrate the difficulty of support tasks to be non-trivial yet attainable through practice.

\td{TD4. Overlapping Skills.} In a co-training paradigm, numerous factors can affect transfer from support tasks to downstream tasks \cite{gao2024efficient, maddukuri2025simandreal, saxena2025matters}. When designing support tasks, we incorporate at least one shared manipulation skill between target and support tasks: for example, grasping objects of similar geometries or inserting items into a small box (\cref{fig:overview}); the specific low-level motion can differ between target and support tasks, but the high-level motion is similar. \looseness=-1 %In service of TD1--3, we vary factors such as object composition, initial distribution, and goal distribution.

\subsection{Task Details}
\label{sec:method.task_details}

We now apply these principles to the design of 3 task classes, shown in \cref{fig:tasks}, and instantiate them in RoboCade. Our task design is guided by 3 target tasks (\tScene, \tScan, \tBox), resulting in 3 support tasks (\sScene, \sScan, \sBox). In \cref{sec:experiments.policy_learning}, we demonstrate that co-training on support tasks can be beneficial for learned policies on target tasks.

\paragraph{Rearrangement (\cref{fig:tasks}; top).} The \tScene target task requires sequentially picking up two pieces of electronics---a USB adapter and a computer mouse, placed in an arbitrary configuration on one side of a desk, alongside a mug---and arranging them in a straight line on the other side of the desk. This task requires  precision in grasping (the adapter lies flat on the table at about 1 cm in height, and the mouse has a curved surface) as well as placement to arrange the items in a line. As a support task, we design the \sScene task: users rearrange two animal blocks  to recreate a given scene (TD1). The goal locations are more diverse (TD2) than in \tScene and are presented to the user with a scene sketch virtually overlaid on the table (SD2).  We vary the specific scene between episodes, and matching the scene exactly with the overlay introduces challenge (TD3). At a high-level, this task requires similar skills of grasping, orienting, and arranging objects as in the \tScene task (TD4). \looseness=-1
        
\paragraph{Scanning (\cref{fig:tasks}; middle).} The \tScan target task involves picking up a bottle of threadlocker solution and orienting the barcode on the bottle in front of a scanning camera. We design a support task \sScan, which is a multi-step task framed as checking out groceries from a receipt within the given time (TD1, SD2). \sScan requires picking up a toy grocery item, bringing it to the scanner, and placing it in a toy grocery basket. Upon successfully scanning, the interface plays an auditory beep, checks off the item from the receipt, and displays celebratory virtual confetti (SD1). This task entails shared skills from \tScan, including grasping and aligning with the scanner (TD4). \sScan also contains the extra step of placing in the grocery basket. To make the task easier, the motion required is simpler in that it does not require a significant amount of reorientation, and grasping the objects is more straightforward than grasping the threadlocker bottle which has a narrow cap (TD3).  For diversity, \sScan uses a variety of objects compared to \tScan which uses just one (TD2). \tScan can be regarded as a specialized version of one skill in \sScan, focusing on a harder object, without the diversity of objects and gamified effects.

\paragraph{Insertion (\cref{fig:tasks}; bottom).} \tBox is a multi-step task which requires picking up a roll of electrical tape, inserting it in a cardboard box, and closing the box shut. This task requires precise motion, particularly when closing the lid along its joint. To provide a support task with a more appropriate level of difficulty, we design the \sBox support task, which involves picking up an animal toy and inserting it into a small toy box of a matching color (TD3). This task is framed as ``returning an animal to its home'' (TD1). This task shares the skills of grasping and insertion with \tBox (TD4). We introduce diversity into this task by (a) varying the animal toy in each episode and (b) randomizing the location of the animal toy and houses across a larger range of initial states than in \tBox (TD2).

\subsection{System Architecture and Implementation Details}

We implement RoboCade using the Franka FR3 robot arm.
For parity with the DROID platform \cite{Khazatsky2024DROIDAL}, we use two third-person ZED2 cameras and one ZED Mini egocentric camera. As the input device, we use GELLO \cite{wu2024gello} with joint impedance control. We select GELLO as it is low-cost, buildable by hobbyists, and allows for intuitive joint-space control, naturally obeying kinematic constraints and avoiding self-collisions; additionally, \cite{wu2024gello} finds that novice users prefer GELLO compared to Cartesian control with a 3D mouse or a VR controller. For safety, we prevent the robot's joints from exiting workspace bounds to avoid collisions with the table and cameras. We use ZeroMQ \cite{zeromq} for message passing and WebSockets for transmitting data between the backend and frontend. Within the main viewer, a live point cloud of objects in the scene is displayed alongside a virtual rendering of the robot as well as additional visual effects. The camera of the 3D viewer moves based on the user's action to provide a sense of depth. We implement the RoboCade frontend using Next.js \cite{nextjs} and Three.js \cite{threejs} for 3D rendering. Additional implementation details are given in the Appendix.
\section{Experiments}
\label{sec:experiments}

In our experiments, we evaluate RoboCade as a data collection platform in two main aspects. First, in \cref{sec:experiments.policy_learning} and \cref{sec:experiments.vla_policy_learning}, we evaluate whether data collected from gamified support tasks can effectively improve downstream policy learning on non-gamified target tasks through co-training. Second, in \cref{sec:experiments.user_study}, we evaluate how usable and engaging the system is via a human subjects study.

\subsection{Policy Learning: Co-training with Gamified Data}
\label{sec:experiments.policy_learning}

\begin{figure*}
    \centering
    \includegraphics[width=\linewidth]{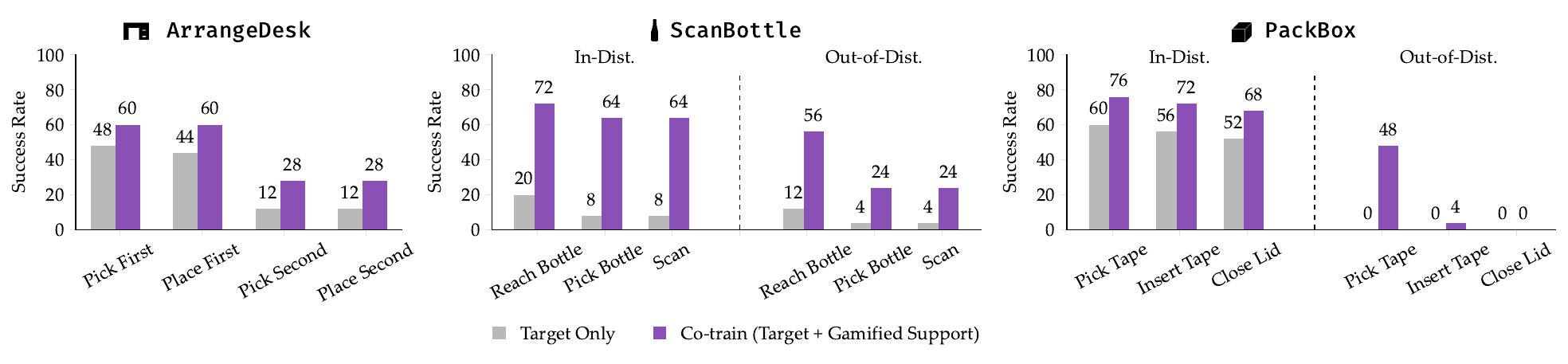}
    \vspace{-5mm}
    \caption{\textbf{Co-training with Gamified Data.} For our 3 target tasks, we compare the performance of Diffusion Policy \cite{chi2023diffusion} when trained only on target task data (\textit{Target Only}) versus co-trained with gamified support tasks (\textit{Co-train}), with a fixed training budget. For each task and training condition, we perform 25 trials and report staged success rate (\%). Co-training improves success rate on all 3 tasks for in-distribution conditions (\textit{In-Dist.}). For \tScan and \tBox, we additionally evaluate on out-of-distribution initial configurations (\textit{Out-of-Dist.}) and find that co-training improves generalization.}
    \label{fig:policy_learning}
    \vspace{-3mm}
\end{figure*}

First, we investigate the usefulness of gamified data in policy learning. We examine how data collected on support tasks can be a viable source of co-training data for target tasks---reducing the reliance on target-specific data.

\paragraph{Experiment Setup.} For each pair of target and support tasks, we collect a pair of datasets ($\mathcal{D}_\tau$, $\mathcal{D}_{\tau'}$). We collect $\mathcal{D}_{\tau}$ through standard in-person teleoperation with GELLO, and $\mathcal{D}_{\tau'}$ through the RoboCade platform. For \tScene and \tScan, $\mathcal{D}_\tau$ consists of 80 demonstrations, and for \tBox, $\mathcal{D}_\tau$ consists of 150 demonstrations. For each support task (\sScene, \sScan, and \sBox), $\mathcal{D}_{\tau'}$ contains 200 demonstrations.
For each task pair, we compare policy performance between training on target data $\mathcal{D}_\tau$ alone to co-training with gamified support data. Training only on $\mathcal{D}_\tau$ constitutes the \textit{Target Only} setting. Co-training on $\mathcal{D}_\tau$ and $\mathcal{D}_{\tau'}$ constitutes the \textit{Co-train} setting.

\paragraph{Training Details.} We train policies with Diffusion Policy \citep{chi2023diffusion} using third-person and egocentric camera images and end-effector position as the observation, and absolute joints as actions. We follow the U-Net architecture from \citep{chi2023diffusion} with action chunks of size 16 and execution horizon of 8. For co-training, we sample batches with a 50\%-50\% split between data sources. For fair comparison, all policies are trained for the same number of steps (300K). Policies operate at 12Hz.

\paragraph{Evaluation Protocol.} For each task and training condition, we perform 25 trials. We randomize the initial poses of objects at each trial, and use image overlays to maintain consistency in initial poses when evaluating different policies on the same task to minimize evaluation variance \cite{lbmtri2025}. We record success rate for each task at different stages.

\paragraph{Results.} \cref{fig:policy_learning} summarizes our results. We find that co-training with gamified data from support tasks improves policy performance across all target tasks, with the magnitude and type of improvement varying by task.

\textbf{\tScene}. This target task has multiple steps: two pick-and-place motions to move the USB adapter and the computer mouse away from the mug and line them up at the opposing side of the table. The relative positions and orientations of the mug, adapter, and mouse are randomized at one side of the table. In the corresponding support task \sScene, two animal blocks are randomized in position and orientation (in the same general region on the table) but their final position varies based on the goal scene. With only 80 demonstrations in $\mathcal{D}_\tau$, the target-only policy for \tScene achieves only 12\% overall success, but co-training with \sScene data increases the success to 28\%. Qualitatively, we observe that successful rollouts are higher quality---i.e., with more aligned grasps and less retrying.

\textbf{\tScan}. In this task, the threadlocker bottle’s initial position and orientation are randomized within a roughly 1.5-square-foot region. In \sScan, the items to be scanned are toy grocery items (e.g., a toy carton of milk). \sScan entails a similar basic motion to \tScan (grasping an object and aligning it in front of the scanner camera), but scanning the threadlocker bottle requires more precision to grasp from the top due to its longer and thinner cap. \sScan involves an extra step of placing the object in a grocery basket. The region of initial object placements in \sScan partially overlaps with that of \tScan, but extends to other areas of the table. Interestingly, we find that co-training with \sScan substantially improves in-distribution performance ($>$50\%) on \tScan, even including regions of the \tScan initial distribution that are not covered by \sScan. We also evaluate \tScan in out-of-distribution initial states---regions that are not covered by \tScan's target data---and find that policy performance improves with co-training by approximately 20\%. We find that co-training with \sScan helps substantially at the stage of localizing the threadlocker bottle precisely enough to grasp.

\textbf{\tBox}. This multi-step task is challenging, requiring first grasping the roll of electrical tape, then inserting it into the box, and lastly closing the lid shut. Thus, we collect a larger set of target data (150 demonstrations). \sBox has shared skills with this task---picking up a small toy and inserting it into a box. \sBox's animal initial positions are more diverse than that of \tBox. We see that co-training with \sBox provides a $16$\% improvement on in-distribution performance, particularly at the stage of picking up the tape roll. Qualitatively, we observe that grasps are more precise after co-training (being aligned closer to the center axis of the roll of tape). We additionally perform out-of-distribution evaluations on \tBox, and find that with co-training, the learned policy \tBox is able to generalize the pick subtask to locations unseen in the \tBox data. Overall, these results validate the idea that gamified support tasks can produce beneficial data for specialized policies of more realistic target tasks.

\subsection{Policy Learning: VLA Co-fine-tuning with Gamified Data}
\label{sec:experiments.vla_policy_learning}

\begin{figure}
    \centering
    \includegraphics[width=\linewidth]{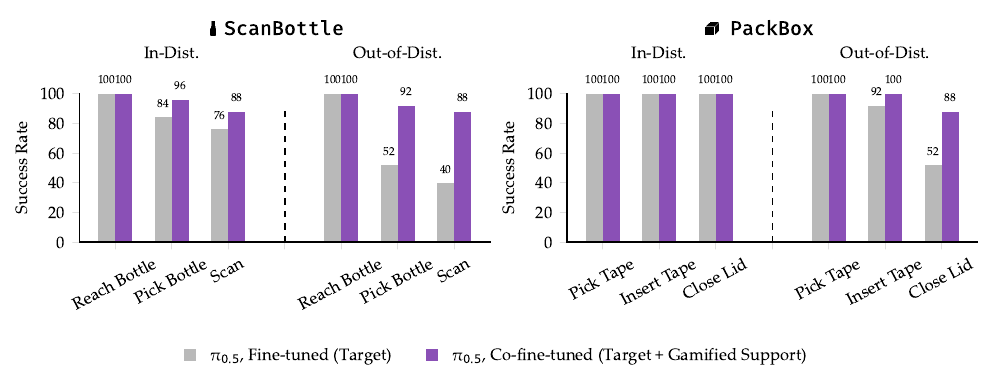}
    \vspace{-5mm}
    \caption{\textbf{VLA Co-fine-tuning with Gamified Data.} We compare the performance of $\pi_{0.5}$ \cite{black2025pi_05} fine-tuned on the DROID dataset \cite{Khazatsky2024DROIDAL} when further fine-tuned on target task data versus co-fine-tuned on both target task and gamified support task data, with a fixed budget of training steps. For each task and training condition, we perform 25 trials. Co-fine-tuning on gamified support task data improves performance in out-of-distribution initial configurations (\textit{Out-of-Dist.}) while matching or exceeding performance in in-distribution conditions (\textit{In-Dist.}). \looseness=-1}
    \label{fig:vla_finetuning_results}
    \vspace{-6mm}
\end{figure}

Next, we examine whether the utility of gamified data holds in the context of pre-trained vision-language-action (VLA) models. \looseness=-1 

\paragraph{Experiment Setup.}
We begin with a version of $\pi_{0.5}$ \cite{black2025pi_05} fine-tuned on the DROID dataset \cite{Khazatsky2024DROIDAL} and further fine-tune it on $\mathcal{D}_{\tau}$ or co-fine-tune it on $\mathcal{D}_\tau$ and $\mathcal{D}_{\tau'}$.
For fair comparison, we perform model selection for each task on the target-only condition to determine the budget of training steps used for both conditions.
Third-person and egocentric camera images are provided as the observation. We use action chunks of size 16 and an execution horizon of 10.
For evaluation, we use the same initial configurations from \cref{sec:experiments.policy_learning} for in-distribution and out-of-distribution conditions, with 25 trials per task and condition.
%This model produces chunk-relative desired joint actions.

\paragraph{Results.} \cref{fig:vla_finetuning_results} summarizes our results. Relative to the policies from \cref{sec:experiments.policy_learning}, both the fine-tuned and co-fine-tuned $\pi_{0.5}$ models achieve substantially higher success rates in both in-distribution and out-of-distribution settings. Both the fine-tuned and co-fine-tuned models can robustly localize and reach the initial object of interaction. Compared to the fine-tuned model, we observe that co-fine-tuning with support task data further improves in-distribution performance in \tScan (+12\%) and maintains performance in \tBox.
Co-fine-tuning also yields gains in generalization to out-of-distribution configurations, improving success on \tScan (+48\%) and \tBox (+36\%). Qualitatively, we find that for \tScan, the co-fine-tuned policy achieves more stable grasps at the picking stage, and for \tBox, the co-fine-tuned policy is more precise at placing the tape in the middle of the box (rather than at the edge) which is necessary to close the lid.
In sum, these results demonstrate that gamified tasks, designed following the principles in \cref{sec:method.task_design}, can provide useful data for downstream policies of target tasks.

\begin{figure}[t!]
    \centering
    \includegraphics[width=\linewidth]{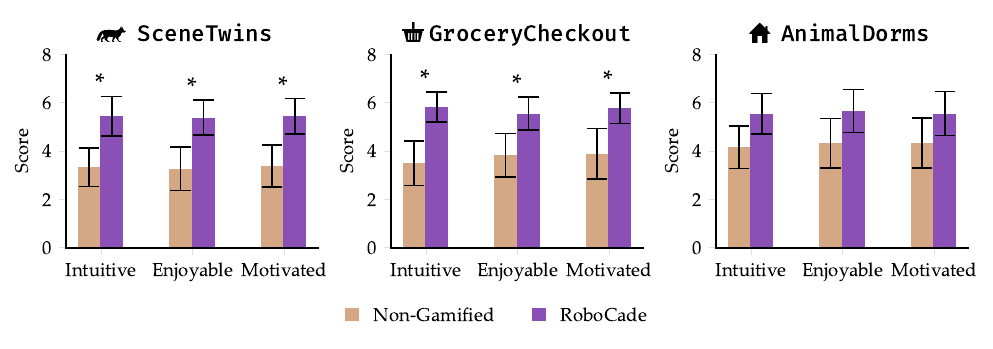}
    % \vspace{-7mm}
    \caption{\textbf{User Study Subjective Results.} We report user rankings for RoboCade compared to a non-gamified remote teleoperation interface. The GELLO controller is used in both conditions. Users tend to find the gamified interface more intuitive, enjoyable, and motivating. Error bars indicate 95\% confidence intervals. * denotes significance at $p<0.05$ under a Wilcoxon signed-rank test. \looseness=-1}
    \label{fig:user_study_likert}
    \vspace{-4mm}
\end{figure}

\subsection{User Study}
\label{sec:experiments.user_study}

Next, we evaluate user engagement and the usability of RoboCade through a human subjects study with $N=18$ novice users coming from diverse backgrounds, including 7 users located in a different U.S.~state than the robot.

\paragraph{Participants and Procedure.} We conduct a within-subjects study with 18 participants (6 female and 12 male, ages 20--36) performing remote teleoperation. For each participant, the study takes approximately 45 minutes to 1 hour.
After obtaining informed consent, each user participates in a familiarization task involving pick-and-place movements, with RoboCade and a baseline remote teleoperation system using the GELLO. The goal of the familiarization task is to allow the user to understand the mechanics of each interface and how to perform teleoperation with the GELLO. Each user is allowed up to 5 minutes of familiarization time with each system.
Next, the user performs 3 tasks (\sScene, \sBox, \sScan). The user performs each task 3 times with both systems, with up to 3 attempts allowed per trial, and a limit of 5 minutes per combination of task and condition.  We counterbalance the order of tasks and conditions. The study is approved under Stanford University's Institutional Review Board. \looseness=-1

\paragraph{Robot Control Method.} We compare RoboCade to a vanilla remote teleoperation interface (a multi-camera view without gamified components). To control for teleoperation input device, we use the same GELLO controller for both conditions.

\paragraph{Dependent Measures.}
After each task, users complete a qualitative survey regarding how intuitive they found each condition, their level of enjoyment, and their level of motivation to continue the session, on a Likert scale from 1--7.

\paragraph{Results.}
We report subjective survey results in \cref{fig:user_study_likert}. We run Wilcoxon signed-rank tests with Holm-Bonferroni correction to compare user ratings for RoboCade and the non-gamified system. Users report that they prefer the gamified RoboCade system as being more intuitive (+27\%), enjoyable (+24\%), and leading to a higher level of motivation (+24\%). This effect is significant for \sScene and \sScan ($p < 0.05$, Holm-corrected). For \sBox, the user preference is only marginally significant, which we hypothesize is because \sBox is a shorter and easier task overall.
Participants also have a higher task completion rate with RoboCade compared to the baseline system (77.1\% versus 69.7\%). Additionally, a paired $t$-test shows that users prefer the gamified system over the baseline in terms of perceived usability, with SUS scores \cite{brooke1996sus} of $71.8$ (SD $20.4$) for the gamified system and 51.4 (SD $22.4$) for the baseline, $t(17) = 3.48$, $p = 0.003$.
\section{Discussion}
\label{sec:discussion}

We propose an approach for scaling robot data collection by designing gamified remote teleoperation experiences that engage general users, expanding the pool of operators to include users motivated by interactive gameplay. Specifically, we introduce RoboCade, a platform that applies gamification principles to both system design and task design. Through policy learning experiments, we demonstrate the viability of using gamified task data to support policy learning and potentially offload some of the cost of collecting target-specific data. Through a human subjects study, we validate the system as an engaging data collection platform, significantly enhancing participant enjoyment compared to a non-gamified baseline.

\paragraph{Limitations and Future Work.} Our current work has limitations that open up several avenues for future work. 

\textit{Data Quality Incentivization.} The general population has largely untapped potential for robot data collection. However, quality control can be challenging in the context of novice users. Future work can extend RoboCade to incorporate quality metrics \cite{zhang2025scizor, hejna2025robot, agia2025cupid}---e.g., through a point system that incentivizes high quality trajectories. Notably, two of our best performing users reported having significant experience playing video games despite a lack of teleoperation experience. This suggests an opportunity to tap into the gaming community for scaling up robot data collection by embedding robot control into longer-form games. \looseness=-1

\textit{In-depth Study of Game Elements.} RoboCade adapts a number of gamification components to robot data collection. While the study of the fine-grained effect of individual game components on users' behavioral and psychological outcomes \cite{koivisto2019rise, landers2017gamification} is outside the scope of this work, our platform can enable more in-depth study of these components in the context of robot data collection.
Future work can also study long-term interactions among general users with the platform  to understand user retention and novelty effects \cite{farzan2008experiment, hamari2014does, koivisto2019rise}, as well as to scale up data collection---while ensuring that users remain informed about how their data is used. \looseness=-1

\textit{Learning from Play.} This work focuses on collecting task-directed interactions and utilizing them for co-training. Another avenue for gamified robot data collection is to endow users with more creativity and agency in their interactions, and collect task-agnostic play data; such data could be used in learning paradigms beyond co-training (e.g., training goal-conditioned methods or world models) \cite{lynch2020learning, cui2023play}. Overall, the intersection of gamification and robotics presents a rich set of research directions which we hope RoboCade can help enable. \looseness=-1

% === Back Matter ===
\section*{Acknowledgements}
We acknowledge funding from Toyota Research Institute, NSF Award 1941722 \& 2006388, ONR YIP \& Grant W911NF2210214. 
We are grateful for discussions with Aaron Tung, Jenn Grannen, Jensen Gao, Shuang Li, Aykut Onol, and Mark Zolotas.
% AI tools (Claude, ChatGPT) were used in a limited capacity (grammar enhancement, code autocompletion). The authors are responsible for all content in this article. 

% \addtolength{\textheight}{-12cm}

\printbibliography

\clearpage

\section*{Appendix}

\subsection{System Implementation Details}
\label{app:implementation_details}

\begin{figure}[h]
    \centering
    \includegraphics[width=\linewidth]{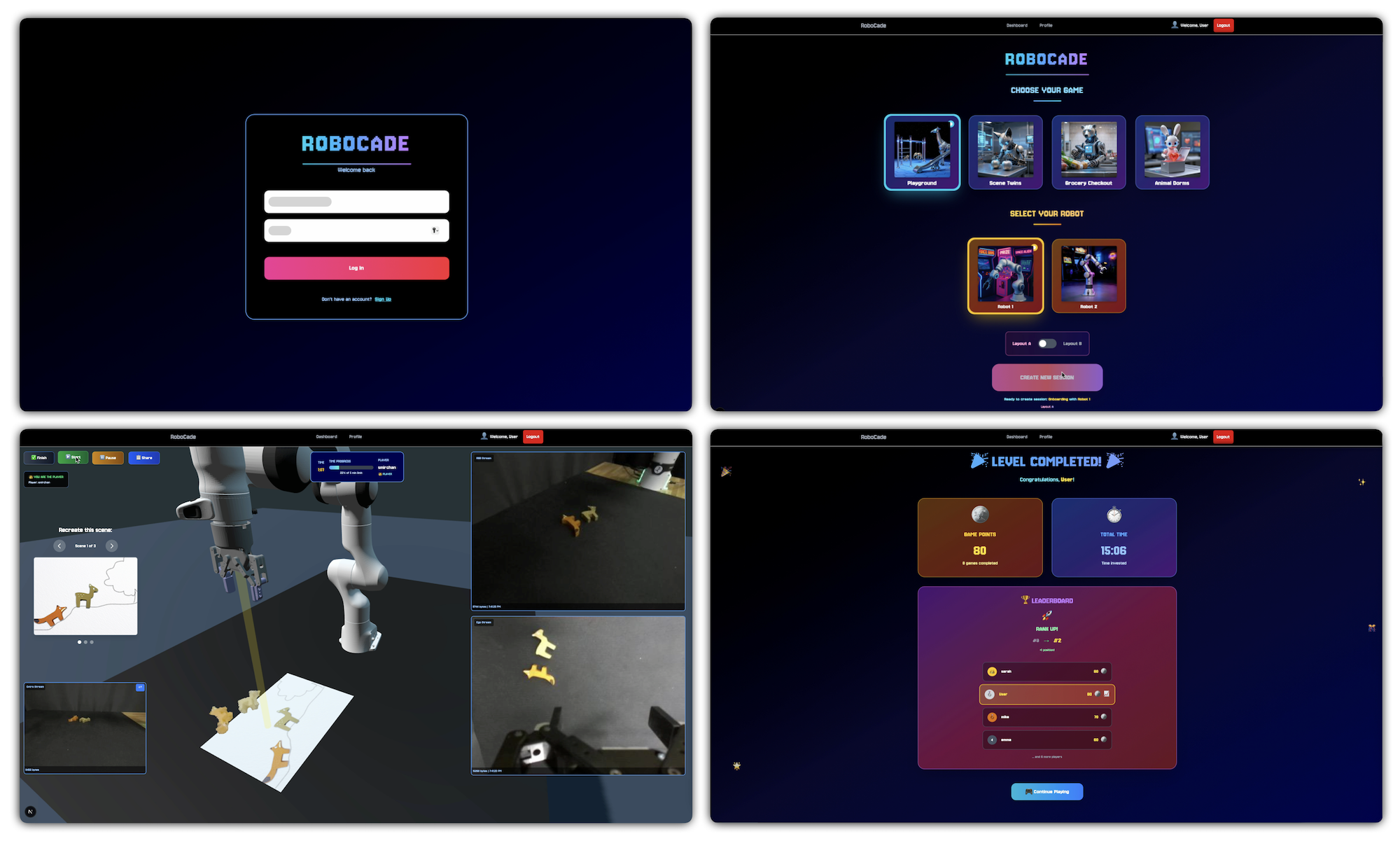}
    \caption{\textbf{Web Interface.} (\textit{Top Left}) Sign-in page. (\textit{Top Right}) Task selection: Users select from available tasks and robots for remote teleoperation. (\textit{Bottom Left}) Task play: While remotely controlling the real robot, users can view a rendering of the robot and objects in the scene. (\textit{Bottom Right}) Post-task: Badges and leaderboard are presented post-task.}
    \label{fig:interface}
\end{figure}

We implement RoboCade with the Franka FR3 robot arm. Similar to the DROID setup \cite{Khazatsky2024DROIDAL}, we mount two third-person ZED2 cameras and one ZED Mini camera on the robot's wrist. We calibrate the external cameras and merge point clouds via point-to-point ICP registration. Compressed point clouds are sent to the frontend and displayed alongside a virtual rendering of the robot via Three.js \cite{threejs}. The camera angle of the 3D viewer moves based on the position of the end-effector to provide the user with a sense of depth, and a virtual beam projects the direction of the end-effector onto the table. We use a publisher-subscriber messaging pattern with ZeroMQ \cite{zeromq} for data processing on the backend, and WebSockets to communicate between the backend and the frontend. The frontend is implemented with Next.js. User accounts and authentication are managed using Supabase. As the input device, we use the Franka FR3 version of the open-source GELLO controller \cite{wu2024gello}. For safety, we perform collision avoidance with the table and workspace boundaries and additionally limit the magnitude of actions during remote teleoperation. We use joint impedance control via Polymetis \cite{polymetis}. We provide screenshots of the web interface in \cref{fig:interface}.

\subsection{Target Task Details}

% Here, we provide details on the target tasks in this work.

\paragraph{\tScene.} This task involves picking up two pieces of electronics (a USB adapter and a computer mouse) placed in an arbitrary configuration on one side of a desk, alongside a mug, and arranging them in a straight line on the other side of the desk. The order, position, and orientation of the mug, adapter, and mouse are randomized in each episode. Picking the first object, placing the first object, picking the second object, and placing the second object in the proper positions constitute the 4 subtasks used to score this task.

\paragraph{\tScan.} This task involves picking up a bottle of threadlocker solution and orienting the barcode in front of a scanning camera. The position and orientation of the bottle are randomized within a roughly 1.5-square-foot region. The following subtasks are used to score this task: reaching the bottle such that the gripper makes contact with the bottle, picking up the bottle from the table, and scanning the bottle such that the correct side of the bottle is at least partly visible by the scanning camera.

\paragraph{\tBox.} This task involves picking up a roll of electrical tape, placing it inside a cardboard box, and closing the box shut. The position of the box and tape are both randomized. The box lid cannot fully close if the tape roll is placed against an edge of the box. The following subtasks are used to score this task: picking up the tape from the table, inserting it into the box, and closing the lid of the box shut.

\subsection{Policy Learning Details}

In this section, we provide hyperparameters for our policy training experiments. Hyperparameters for Diffusion Policy training are given in \cref{tab:diffusion_hyperparams} and for VLA fine-tuning are given in \cref{tab:vla_hyperparams}. For fair comparison, we use a fixed number of training steps for training and co-training policies (300K for Diffusion Policy). For $\pi_{0.5}$ fine-tuning, we perform model selection among 5K, 10K, and 20K training steps with real-world evaluation on the target-only fine-tuned policy and use the same number of training steps for co-fine-tuning.

\begin{table}[ht]
    \centering
    \small
    \begin{tabular}{c c}
        \toprule
        Training Steps          & 300K \\
        Batch Size              & 128 \\
        Optimizer               & AdamW \\
        $\beta_1$               & 0.95 \\
        $\beta_2$               & 0.999 \\
        Learning Rate           & 1e-4 \\
        Weight Decay            & 1e-6 \\
        LR Schedule             & Cosine Decay \\
        Warmup Steps     & 2000 \\
        Diffusion Architecture  & Conv1D UNet \\
        Prediction Horizon      & 16 \\
        Observation History     & 1 \\
        Num Action              & 8 \\
        % Kernel Size             & 5 \\
        % Num Groups              & 8 \\
        Step Embedding Dim       & 128 \\
        UNet Down Dims           & [256, 512, 1024] \\
        Num Diffusion Steps     & 100 \\
        Num Inference Steps     & 20 \\
        Inference Scheduler     & DDIM \\
        % Observation Input       & FiLM \\
        % Image Encoder           & ResNet-18 \\
        Image Embedding Dim     & 256 \\
        % Proprioception          & Yes \\
        Cameras                 & Third Person, Wrist \\
        \bottomrule
    \end{tabular}
    \caption{Hyperparameters for Diffusion Policy}
    \label{tab:diffusion_hyperparams}
\end{table}

\begin{table}[ht]
    \centering
    \small
    \begin{tabular}{c c}
        \toprule
        Base Model              & $\pi_{0.5}$-DROID \\
        Batch Size              & 32 \\
        Optimizer               & AdamW \\
        $\beta_1$               & 0.9 \\
        $\beta_2$               & 0.95 \\
        Learning Rate           & 2.5e-5 \\
        Gradient Clip Norm      & 1.0 \\
        LR Schedule             & Cosine Decay \\
        Warmup Steps            & 1000 \\
        Prediction Horizon      & 16 \\
        Num Action              & 10 \\
        % Proprioception          & Yes \\
        Cameras                 & Third Person, Wrist \\
        \bottomrule
    \end{tabular}
    \caption{Hyperparameters for $\pi_{0.5}$ Fine-tuning}
    \label{tab:vla_hyperparams}
\end{table}

\end{document}